\documentclass[sigconf]{acmart}

\usepackage{multirow}
\usepackage{makecell}
\usepackage{tabularx}
\usepackage{pifont}
\usepackage{xspace}
\usepackage{algorithm}
\usepackage{algpseudocode}

\copyrightyear{2026}
\acmYear{2026}
\setcopyright{cc}
\setcctype{by}
\acmConference[SIGSPATIAL '26]{The 34th ACM International Conference on Advances in Geographic Information Systems}{November 03--06, 2026}{Riverside, CA, USA}
\acmBooktitle{The 34th ACM International Conference on Advances in Geographic Information Systems (SIGSPATIAL '26), November 03--06, 2026, Riverside, CA, USA}
\acmDOI{10.1145/3841645.3843433}
\acmISBN{979-8-4007-2950-8/2026/11}

\begin{document}

\title{Unlearning on Spatio-Temporal Graphs through Subgraph Virtual Edge Reconstruction}

\author{Qiming Guo}
\correspondingauthor
\affiliation{%
  \institution{Texas A\&M University - Corpus Christi}
  \city{Corpus Christi}
  \state{Texas}
  \country{USA}
}
\email{qguo2@islander.tamucc.edu}

\author{Wenbo Sun}
\affiliation{%
  \institution{Delft University of Technology}
  \city{Delft}
  \country{Netherlands}
}
\email{w.sun-2@tudelft.nl}

\author{Chen Pan}
\affiliation{%
  \institution{University of Texas at San Antonio}
  \city{San Antonio}
  \state{Texas}
  \country{USA}
}
\email{chen.pan@utsa.edu}

\author{Ye Wang}
\affiliation{%
  \institution{Biogen}
  \city{Cambridge}
  \state{Massachusetts}
  \country{USA}
}
\email{ye.wang@biogen.com}

\author{Wenlu Wang}
\correspondingauthor
\affiliation{%
  \institution{Texas A\&M University - Corpus Christi}
  \city{Corpus Christi}
  \state{Texas}
  \country{USA}
}
\email{wenlu.wang@tamucc.edu}

\renewcommand{\shortauthors}{Guo et al.}

\begin{abstract}
Spatio-temporal graphs are widely used in modeling complex dynamic processes such as temporal forecasting, molecular dynamics, and healthcare monitoring. Recently, stringent privacy regulations such as GDPR and CCPA have introduced significant new challenges for existing spatio-temporal graph models, requiring complete unlearning of unauthorized data. Since each node in a spatio-temporal graph diffuses information globally across both spatial and temporal dimensions, existing unlearning methods primarily designed for static graphs and localized data removal cannot efficiently erase a single node without incurring costs nearly equivalent to full model retraining. To address this, we propose CallosumNet, a spatio-temporal graph unlearning framework biologically inspired by the corpus callosum structure. CallosumNet makes two key technical contributions: (1)~it reconstructs subgraphs using \textbf{biologically-inspired virtual edges}; and (2)~it restores \textbf{interlinked spatio-temporal dependencies} among subgraphs via a lightweight meta-graph integration layer. Empirical results on four diverse real-world datasets show that CallosumNet achieves complete unlearning while maintaining accuracy very close to the gold model. The code is publicly available at \url{https://github.com/wenlu-lab/STGraphUnlearning}.
\end{abstract}



\maketitle

\begin{figure}[!b]
    \centering
    \includegraphics[width=\linewidth]{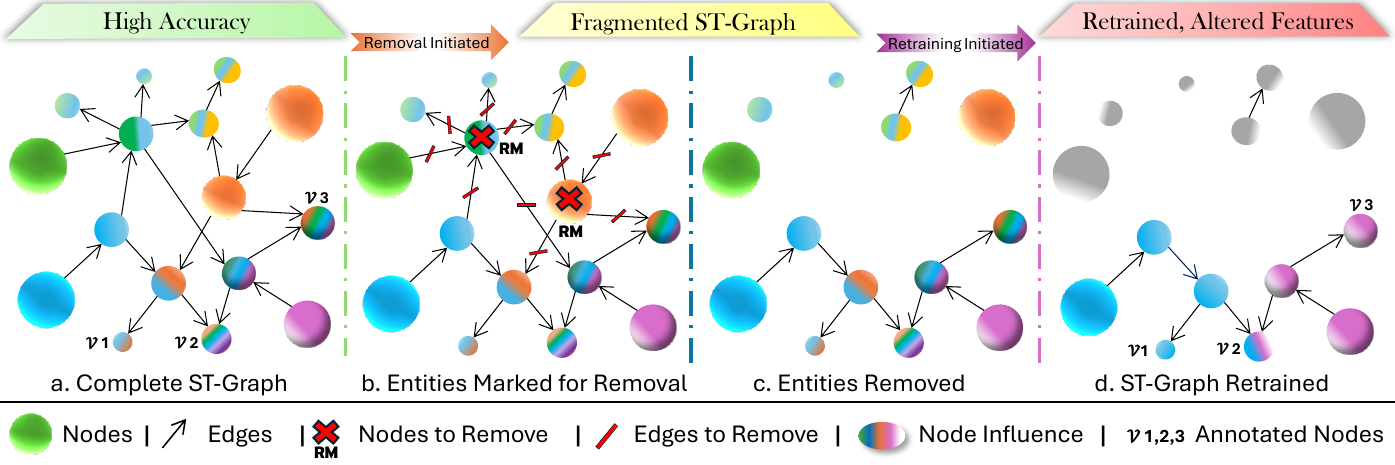}
    \caption{Unlearning on an ST-graph: (a) full graph; (b) consent-revoked nodes marked; (c) record deletion leaves residual model influence; (d) retraining purges it but fragments the graph and distorts remaining features ($v_1$--$v_3$).}
    \Description{Four panels showing a spatio-temporal graph before deletion, with nodes marked for removal, after raw record deletion with residual influence, and after full retraining with fragmented structure and altered node features.}
    \label{figure1}
\end{figure}

\section{Introduction}

Recent advanced spatio-temporal graph models effectively capture complex dynamic processes, such as urban traffic flows, molecular interactions, and healthcare monitoring, by harnessing both spatial adjacency and temporal continuity. However, the broad deployment of these powerful models increasingly faces stringent privacy regulations, such as the General Data Protection Regulation (GDPR) and the California Consumer Privacy Act (CCPA), which necessitate the complete removal or \textit{unlearning} of sensitive user data upon request.

\begin{figure*}[!t]
    \centering
    \includegraphics[width=0.7\linewidth]{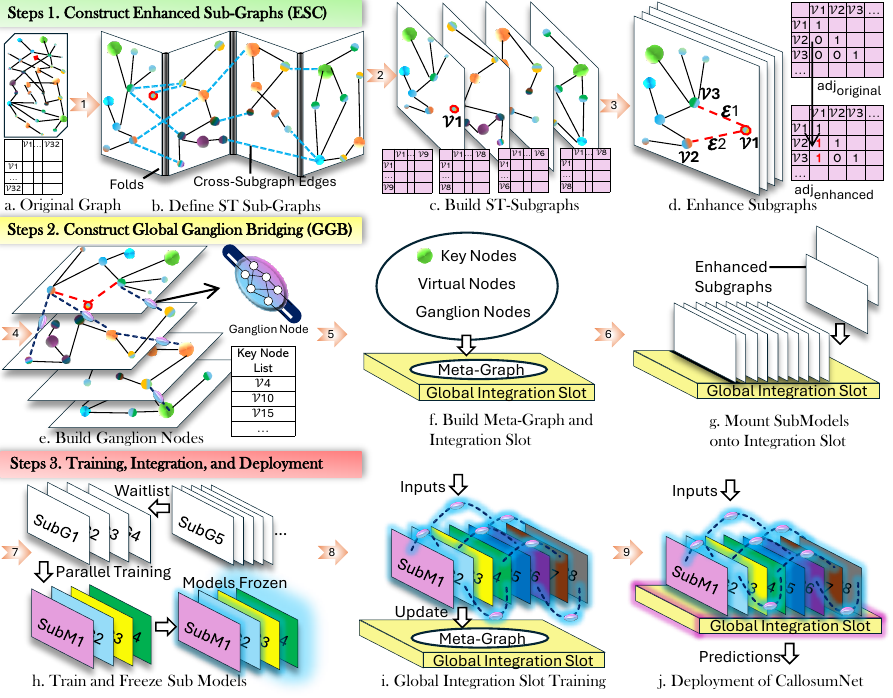}
    \caption{CallosumNet system construction.
    The original graph (a) is transformed into multiple enhanced local subgraphs (d) through ESC, and then the GGB method adds ganglion nodes and identifies key nodes to construct the meta-graph.}
    \Description{Nine-step pipeline diagram showing how the original graph is partitioned into enhanced subgraphs via ESC, how ganglion and key nodes form the meta-graph and global integration slot via GGB, and how sub-models are trained, frozen, integrated, and deployed.}
    \label{figure2}
\end{figure*}

\paragraph{Motivating scenario.}
Taking a mobile–location service (e.g., Google Maps) as an example, Figure~\ref{figure1}(a) shows smartphones (nodes) forming a richly coupled spatio-temporal graph stream of time-stamped GPS signals. Suppose a subset of users revokes consent for their location data, necessitating the deletion of these devices and all incident edges, as shown in Figure~\ref{figure1}(b). Simply dropping the raw records (Figure~\ref{figure1}(c)) does not fully satisfy the deletion requirement, as it fails to eliminate the latent influence of the revoked users. Conversely, retraining the entire model from scratch after purging those records (Figure~\ref{figure1}(d)) erases the influence but fragments long-range spatial and temporal paths, severely degrading accuracy and interpretability for the remaining users, with a prohibitively high retraining cost.

In this study, we propose CallosumNet, inspired by the corpus callosum, a bundle of $\sim$2$\times$10$^{8}$ axonal fibers that enables the two cerebral hemispheres to specialize independently while staying synchronized~\citep{aboitiz1992fiber}. CallosumNet mirrors this organization through \emph{subgraph virtual edge reconstruction}: it partitions the ST-graph into locally coherent subgraphs, each trained exclusively on its own data like an independent hemisphere, and then reconstructs the severed cross-partition dependencies through biologically-inspired virtual edges and a lightweight meta-graph integration layer---analogous to the corpus callosum bridging the two hemispheres. This design simultaneously enables exact unlearning (each node's influence is confined to one subgraph) and preserves predictive accuracy (global context is recovered without sharing training data across subgraphs). 


\section{Related Work}
\label{sec:preliminary}

Graph unlearning methods mainly fall into approximate and partition-based approaches. Approximate methods, such as influence functions~\citep{koh2017influence} and  GNNDelete~\citep{cheng2023gnndelete}, avoid full retraining but provide only approximate removal and do not explicitly handle spatio-temporal dependencies. Partition-based methods, including SISA~\citep{bourtoule2021machine}, GraphEraser~\citep{chen2022graph}, GraphRevoker~\citep{zhang2025dynamic}, and STEPs~\citep{Guo_Pan_Zhang_Wang_2025}, enable efficient or exact unlearning by retraining only affected partitions, but partition boundaries can break spatial and temporal dependencies, while fixed aggregation cannot fully recover the lost context. CallosumNet addresses this gap by preserving cross-partition dependencies with virtual ganglion edges and restoring global context through a learnable bridging layer.

\section{Methodology}
\label{sec:methods}

We propose \textbf{CallosumNet} (Figure~\ref{figure2}), a partition-and-integrate framework for spatio-temporal graph unlearning. Given an ST-GNN trained on a graph $\mathcal{G}'=(\mathcal{V}',\mathcal{E}',\mathbf{X}')$ with features $\mathbf{X}'\in\mathbb{R}^{T\times N'\times F}$ over $T$ steps, and a deletion request $\mathcal{U}=(\mathcal{U}_N,\mathcal{U}_E)$ of nodes and edges to erase, the goal is a model that behaves as if $\mathcal{U}$ had never been trained on. CallosumNet consists of two components: \textit{Enhanced Subgraph Construction (ESC)} for graph decomposition, and \textit{Global Ganglion Bridging (GGB)} to restore global coherence, organized in a three-step pipeline.

\textbf{1. Divide (ESC).}  Enhanced Sub-graph Construction slices the original ST-graph into $M$ locally coherent sub-graphs along a correlation-driven backbone and patches every cut with virtual ganglion edges so that high-order spatial–temporal paths are preserved. \textbf{2. Link (GGB).}  Global Ganglion Bridging then assembles the sub-graphs into a lightweight meta-graph: it promotes the top-$K$ key nodes, the interface boundary nodes, and the newly created ganglion nodes to meta-graph vertices and sparsely wires them together. Each sub-graph is trained independently (and can be frozen afterwards). Their embeddings are routed through a cross-fusion Transformer that sits on the meta-graph layer and outputs the final prediction. \textbf{3. Unlearning on demand.} When a deletion request arrives, only the sub-graphs that contain the target nodes/edges are re-trained; the meta-graph parameters are fine-tuned, while untouched sub-graphs remain frozen.

\noindent\underline{\textbf{1. Enhanced Subgraph Construction (ESC)}} decomposes $\mathcal{G}'$ into $M$ localized subgraphs while maintaining global dependencies through virtual ganglion edges. For each directed edge $(u,v) \in \mathcal{E}'$,
we compute a $W$-step temporal correlation
$
\rho(u,v) \;=\; \frac{1}{W}\sum_{t=1}^{W}
        \text{corr}\!\bigl(X'_{t,u}, X'_{t+1,v}\bigr),
$
and extract a backbone path
$\mathcal D = \arg\max_{\mathcal P} \sum_{(u,v)\in\mathcal P}\rho(u,v)$,
where $\mathcal{P}$ ranges over all Hamiltonian paths on $\mathcal{V}'$.
Since this maximization is NP-hard, we solve it approximately via a greedy algorithm that iteratively appends the highest-correlation neighbor. Nodes are assigned to subgraphs according to their backbone index:
$
\mathcal V_i
  = \bigl\{\,v \!\in\! \mathcal D \,\bigl\lvert\,
      \lfloor(i{-}1)\tfrac{N'}{M}\rfloor
      \le \text{idx}(v) < \lfloor i\tfrac{N'}{M}\rfloor \bigr\},
$
where $N'=|\mathcal{V}'|$.
Edges internal to $\mathcal V_i$ form $\mathbf A_i$;
the remainder are the cut set $\mathcal{E}_{\text{cut}}$.
Isolated vertices are re-connected to their two nearest neighbours on
$\mathcal D$, and for every $(u,v)\in\mathcal{E}_{\text{cut}}$
we insert a virtual ganglion edge to preserve high-order dependencies.

Additionally, ESC applies a \emph{K-Ring} augmentation within each subgraph: boundary nodes (those incident to cut edges) are sorted by their angular positions in a spring-layout embedding and connected sequentially into a closed ring, so that boundary nodes---most affected by partitioning---stay mutually connected.
The number of partitions is chosen by
$
M^{*} \;=\;
\arg\min_{M}\Bigl[\,
      \Delta_{\text{cut}} + \gamma\log M
    \Bigr]
$,
$
\Delta_{\text{cut}}
  = \!\!\sum_{(u,v)\in\mathcal{E}_{\text{cut}}}\!\!\rho(u,v),
$
with $\gamma$ balancing correlation loss against model parallelism.

\textit{Theorem 1.}\label{theorem:1} Minimising $\Delta_{\text{cut}}$ under equal-size constraints is NP-hard, yet the greedy backbone yields a $(1-\tfrac1e)$ approximation.

\textit{Theorem 2.}
\label{theorem:esc_time_complexity}
ESC runs in $O\!\bigl(T|\mathcal{E}'| + N'^{2}/M\bigr)$ time and stores
$O(N'^{2}/M)$ edges, which is sub-linear in $N'$ when
$M=\Theta(\sqrt{N'})$.
Moreover it retains at least
$\text{Info}_{\text{intra}}
  \ge \bigl(1-\frac{\Delta_{\text{cut}}}{\text{TotalCorr}}\bigr)
        \text{TotalCorr}$
of the total temporal correlation.

\paragraph{Proof sketch.}
Virtual ganglion edges reconnect every isolated vertex to a neighbor with $A'[u,v]>0$ (which exists since $\mathcal{G}'$ is connected), and since $\mathcal{E}'=\bigcup_i\mathcal{E}_i\cup\mathcal{E}_{\text{cut}}$, the retained correlation is exactly $\text{TotalCorr}-\Delta_{\text{cut}}$; a balanced-cut lower bound $\Delta_{\text{cut}}\ge\frac{c}{M}\text{diam}(\mathcal{G}')$ shows the greedy backbone is near-optimal.

\begin{figure*}[h!]
    \centering
    \includegraphics[width=0.8\linewidth]{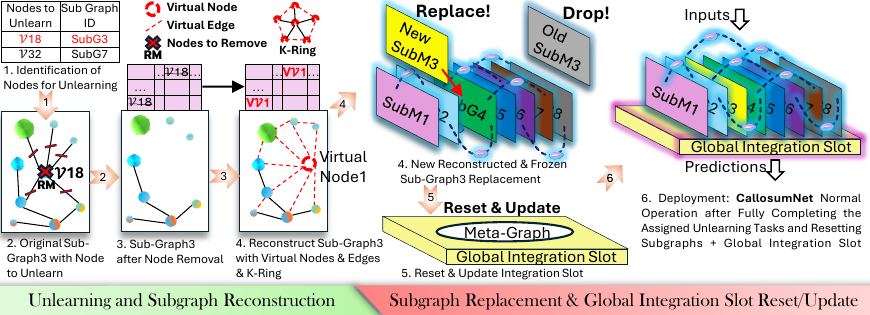}
    \caption{CallosumNet unlearning process}
    \Description{Six-step diagram of the unlearning process: identifying nodes to unlearn, removing them from the host subgraph, reconstructing the subgraph with virtual nodes, edges, and K-Ring, replacing the frozen sub-model, resetting and updating the global integration slot, and redeploying CallosumNet.}
    \label{unlearning}
\end{figure*}

\noindent\underline{\textbf{2. Global Ganglion Bridging (GGB)}} reconstructs global spatio-temporal dependencies by stitching the
$M$ sub-graphs into a lightweight meta-graph
$\mathcal{M} = (\mathcal{V}_{\text{meta}}, \mathcal{E}_{\text{meta}})$
with adjacency matrix $\mathbf{A}_{\text{meta}}$.
It integrates three types of vertices:
(i) \emph{key nodes} (top‐$K$ PageRank per sub-graph, $K=\lceil\log|\mathcal V_i|\rceil$),
(ii) \emph{boundary nodes} incident to cut edges, and
(iii) \emph{ganglion nodes}, each parameterised by a two-layer MLP with ReLU.

Let $\mathcal{E}_{\text{agg}}$ denote the original cross-partition edges (i.e., edges in $\mathcal{E}_{\text{cut}}$ that now connect boundary nodes across subgraphs), and $\mathcal{E}_{\text{key}}$ the edges among key nodes within the same subgraph.
The meta-graph edges are then defined as
$
\mathcal{E}_{\text{meta}}
  = \mathcal{E}_{\text{agg}}
    \cup
    \bigl\{ (u,g),(g,v)
            \mid g\!\in\!\mathcal{V}_{\text{ganglion}},
                 u,v\!\in\!\mathcal{V}_{\text{key}}
                       \cup\mathcal{V}_{\text{boundary}}
      \bigr\}
    \cup
    \mathcal{E}_{\text{key}},
$
and are sparsified until
$\lvert\mathcal{E}_{\text{meta}}\rvert \approx O(M\log M)$.
Each sub-graph is encoded by a frozen STGCN~\citep{stgcn}
$h_v = \text{STGCN}(X'[:,v,:], \mathbf A_i)$
optimised via

$\mathcal{L}_{\text{sub}} = \sum_{v \in \mathcal{V}_i \setminus \mathcal{U}} \bigl\| y_v - \text{pred}_{S_i}(v) \bigr\|_2^{2} + \lambda_{\text{reg}}\lVert\theta_i\rVert_2^{2}$,

thereby isolating $\mathcal U$.
Token-level outputs and ganglion embeddings are fused through a
cross-attention Transformer:
$
h^{\text{final}}
  = \alpha h^{\text{tok}} + (1-\alpha)h^{\text{gang}}
$,

$\hat{y}_v = \text{Transformer}\bigl(\{h'_u,h_g\},\,\mathbf{A}_{\text{meta}}\bigr)$,

where $\alpha$ is a learnable scalar initialised to $0.5$ and clipped to
$[0,1]$.  The overall loss is
$
\mathcal{L}_{\text{ggb}}
  = \sum_{v}\lVert y_v-\hat{y}_v\rVert_2^{2}
    + \lambda_{1}\lVert\mathbf{A}_{\text{meta}}\rVert_{1}
    + \lambda_{2}\sum_{g}\lVert h_g\rVert_2^{2},
    \label{eq:ggb_loss}
$
where $\lambda_{1},\lambda_{2}\ge 0$ control sparsity and embedding regularization, respectively.

\textit{Theorem 3 (Prediction error bound).} For a graph $\mathcal G'$ partitioned into $M$ sub-graphs,
\begin{equation*}
    \bigl\|\hat{y}_{\text{full}} - \hat{y}_{\text{GGB}}\bigr\|_{2}
      \;\le\;
      C_{\text{approx}}\,
        \frac{\Delta_{\text{cut}}\sqrt{M}}{H\,L\,D_g}
\end{equation*}
where $H$, $L$, $D_g$ are the fusion Transformer's head count, depth, and ganglion width, and $C_{\text{approx}}$ is an architecture-dependent constant (see proof sketch).
This bound stays below $0.05$ whenever $M\!\le\!16$ and $N'\!\le\!10^{4}$.

\textit{Theorem 4 (Unlearning stability).} After erasing an arbitrary set $\mathcal U$,
\begin{equation*}
\mathbb{E}\!\bigl[
  \lVert \hat{y}_v - \hat{y}_v^{\text{unlearn}}\rVert_2^{2}
  \;\bigl|\;
  v \notin \mathcal U
\bigr]
\;\le\;
\frac{\Delta_{\text{cut}}\lvert\mathcal U\rvert}
    {(\lvert\mathcal V'\rvert-\lvert\mathcal U\rvert)\,H\,L\,D_g}
\end{equation*}
and the fine-tune converges to an $\varepsilon$-accurate solution, $\varepsilon = G^{2}/(2\eta\sqrt{T_{\text{ep}}})$, for gradient bound $G$, learning rate $\eta$, and $T_{\text{ep}}$ fine-tuning epochs.

\textit{Theorem 5 (Model complexity).} GGB contributes $\mathcal{O}(M\log M\,D_g^{2})$ additional parameters on top
of the $\mathcal{O}(N d^{2}/M)$ parameters of the sub-graphs,
and its per-batch FLOPs are
$\mathcal{O}\!\bigl(BT\,[\,|\mathcal E|/M + M\log M\,]\,d\bigr)$.
With $M=\sqrt{N}$ this yields a sub-linear ($\approx1/\sqrt N$) speed-up compared to a full-graph ST-GNN.

\paragraph{Proof sketch.}
Theorem~3 follows from the Transformer's universal approximation property~\citep{yun2020}; Theorem~4's rate is standard Adam analysis under bounded gradients; Theorem~5 counts $O(d^2 N/M)$ parameters per sub-graph and $O(M\log M\,D_g^2)$ for the meta-Transformer, so $M=\sqrt{N}$ gives $O(\sqrt{N}d^2)$ total.

\noindent\underline{\textbf{3. Unlearning On Demand}} As shown in Figure~\ref{unlearning}, upon receiving a deletion request $\mathcal{U}$, CallosumNet executes three steps: \textit{(i) Locate and remove.} The target nodes and edges are identified and zeroed out in their host subgraph(s); untouched subgraphs remain frozen. \textit{(ii) Reconstruct and retrain.}
ESC re-enhances the affected subgraph(s) with virtual ganglion edges and K-Ring connections, then retrains and freezes them. \textit{(iii) Reset GGB.} The ganglion MLPs and cross-fusion Transformer are reinitialized and fine-tuned (1--3 epochs) on all subgraph outputs, completing the erasure. Since each sub-model trains \emph{exclusively} on its own partition, $\mathcal{U}$'s training signal resides only in its host subgraph(s) and the GGB layer; retraining both leaves the model identical in distribution to one retrained from scratch, i.e., $I(\hat{y};\mathcal{U})=0$, by the same shard-isolation argument as SISA~\citep{bourtoule2021machine}.

\begin{table}[b!]
\centering
\caption{Average relative MAE degradation (\%) versus the gold model, averaged over four backbones (STGCN~\cite{stgcn}, ST-GAT~\cite{velickovic2018graph}, ST-GATV2~\cite{brody2022attentive}, ST-SAGE~\cite{hamilton2017inductive}). Negative values indicate that the method outperforms the gold model.}
\label{tab:rel_delta}
\small
\setlength{\tabcolsep}{4pt}
\renewcommand\arraystretch{1.15}
\begin{tabular}{@{}ll rrrr r@{}}
\toprule
\textbf{Dataset} & \textbf{Rate} & \textbf{SISA} & \textbf{STEPs} & \textbf{GE} & \textbf{GR} & \textbf{Ours} \\
\midrule
\textsc{RWW} & 0\% & +65.3\% & +276.4\% & +734.0\% & +725.8\% & \textbf{+0.0\%} \\
 & 10\% & +57.1\% & +302.4\% & +708.9\% & +664.8\% & \textbf{$-$3.2\%} \\
\midrule
\textsc{PeMS08} & 0\% & +19.1\% & +183.9\% & +107.5\% & +216.1\% & \textbf{+1.1\%} \\
 & 10\% & +13.7\% & +211.6\% & +99.0\% & +211.2\% & \textbf{+1.0\%} \\
\midrule
\textsc{Weather} & 0\% & +9.6\% & +55.1\% & +47.5\% & +66.8\% & \textbf{+4.2\%} \\
 & 10\% & +10.6\% & +55.3\% & +49.3\% & +66.1\% & \textbf{+5.1\%} \\
\midrule
\textsc{Mobility} & 0\% & +27.4\% & +150.5\% & +75.0\% & +251.8\% & \textbf{+7.2\%} \\
 & 10\% & +28.2\% & +152.8\% & +81.3\% & +231.7\% & \textbf{+14.4\%} \\
\midrule
\multicolumn{2}{l}{\textbf{Overall Avg}} & +28.9\% & +173.5\% & +237.8\% & +304.3\% & \textbf{+3.7\%} \\
\bottomrule
\multicolumn{7}{@{}l@{}}{\scriptsize GE = GraphEraser, GR = GraphRevoker, Ours = CallosumNet.}
\end{tabular}
\end{table}

\section{Evaluation}
\label{sec:experiments}

We evaluate on four spatio-temporal graph datasets (23--3{,}220 nodes): RWW~\citep{guo2024hydronet}, PeMS08~\citep{j49q-ch56-25}, Global Weather~\citep{noaa_psl_2025}, and Human Mobility Flow~\citep{kang2020multiscale}, against SISA~\citep{bourtoule2021machine}, STEPs~\citep{Guo_Pan_Zhang_Wang_2025}, GraphEraser~\citep{chen2022graph}, and GraphRevoker~\citep{zhang2025dynamic}. The \emph{gold model} retrains the backbone from scratch on the post-deletion graph and thus carries zero residual influence; \emph{Rate}~$r$ is the fraction of nodes deleted uniformly at random (5 fixed seeds), so $r{=}0\%$ isolates partitioning overhead and $r{=}10\%$ measures post-unlearning accuracy. As Table~\ref{tab:rel_delta} summarizes, CallosumNet degrades MAE by only 3.7\% on average, versus 28.9\% for the best baseline (SISA) and over 170\% for the rest; on Mobility ($N{=}3{,}220$) it also cuts unlearning time from 12{,}640\,s (full retraining) to 3{,}631\,s at $M{=}16$.

\section{Conclusion}
CallosumNet delivers exact spatio-temporal graph unlearning with near-gold accuracy and sub-linear unlearning cost. Limitations include the static-adjacency assumption and meta-graph overhead at large $M$; future work targets dynamic topologies and edge-/feature-level deletion requests.

\begin{acks}
This work was partially supported by the NSF award No.2318641 and No.2112631.
\end{acks}

\bibliographystyle{ACM-Reference-Format}
\bibliography{references}

@inproceedings{cheng2023gnndelete,
  title     = {{GNNDelete}: A General Strategy for Unlearning in Graph Neural Networks},
  author    = {Cheng, Jiali and Dasoulas, George and He, Huan and Agarwal, Chirag and Zitnik, Marinka},
  booktitle = {International Conference on Learning Representations (ICLR)},
  year      = {2023}
}

@article{stgcn,
  author = {Yu, Bing and Yin, Haoteng and Zhu, Zhanxing},
  title = {Spatio-Temporal Graph Convolutional Networks: A Deep Learning Framework for Traffic Forecasting},
  journal = {International Joint Conference on Artificial Intelligence (IJCAI)},
  year = {2018},
  pages = {3634--3640},
  doi = {10.24963/ijcai.2018/505}
}

@article{transformer,
  author = {Vaswani, Ashish and Shazeer, Noam and Parmar, Niki and Uszoreit, Jakob and Jones, Llion and Gomez, Aidan N. and Kaiser, Łukasz and Polosukhin, Illia},
  title = {Attention is All You Need},
  journal = {Advances in Neural Information Processing Systems (NeurIPS)},
  year = {2017},
  volume = {30},
  pages = {5998--6008},
  note = {Available at \url{https://arxiv.org/abs/1706.03762}}
}

@article{yun2020,
  author = {Yun, Chulhee and Bhojanapalli, Sashank and Rawat, Ankit Singh and Reddi, Sashank J. and Kumar, Sanjiv},
  title = {Are Transformers Universal Approximators of Sequence-to-Sequence Functions?},
  journal = {International Conference on Learning Representations},
  year = {2020},
  url = {https://openreview.net/pdf?id=ByxZX0KtDr}
}

@inproceedings{bourtoule2021machine,
  title={Machine unlearning},
  author={Bourtoule, Lucas and Chandrasekaran, Varun and Choquette-Choo, Christopher A and Jia, Hengrui and Travers, Adelin and Zhang, Baiwu and Lie, David and Papernot, Nicolas},
  booktitle={2021 IEEE symposium on security and privacy (SP)},
  pages={141--159},
  year={2021},
  organization={IEEE}
}

@inproceedings{chen2022graph,
  title={Graph unlearning},
  author={Chen, Min and Zhang, Zhikun and Wang, Tianhao and Backes, Michael and Humbert, Mathias and Zhang, Yang},
  booktitle={Proceedings of the 2022 ACM SIGSAC conference on computer and communications security},
  pages={499--513},
  year={2022}
}

@article{Guo_Pan_Zhang_Wang_2025, title={Efficient Unlearning for Spatio-temporal Graph (Student Abstract)}, volume={39}, url={https://ojs.aaai.org/index.php/AAAI/article/view/35259}, DOI={10.1609/aaai.v39i28.35259}, number={28}, journal={Proceedings of the AAAI Conference on Artificial Intelligence}, author={Guo, Qiming and Pan, Chen and Zhang, Hua and Wang, Wenlu}, year={2025}, month={Apr.}, pages={29382-29384} }

@inproceedings{zhang2025dynamic,
  title={Dynamic Graph Unlearning: A General and Efficient Post-Processing Method via Gradient Transformation},
  author={Zhang, He and Wu, Bang and Yang, Xiangwen and Yuan, Xingliang and Liu, Xiaoning and Yi, Xun},
  booktitle={Proceedings of the ACM on Web Conference 2025},
  pages={931--944},
  year={2025}
}

@inproceedings{guo2024hydronet,
  title={HydroNet: A Spatio-temporal Graph Neural Network for Modeling Hydraulic Dependencies in Urban Wastewater Systems},
  author={Guo, Qiming and Wang, Wenlu},
  booktitle={Proceedings of the 32nd ACM International Conference on Advances in Geographic Information Systems},
  pages={717--718},
  year={2024}
}

@article{kang2020multiscale,
  title     = {Multiscale Dynamic Human Mobility Flow Dataset in the U.S. during the COVID-19 Epidemic},
  author    = {Kang, Yuhao and Gao, Song and Liang, Yunlei and Li, Mingxiao and Kruse, Jake},
  journal   = {Scientific Data},
  volumn    = {7},
  issue     = {390},
  pages     = {1--13},
  year = {2020}
}

@misc{noaa_psl_2025,
  author = {{NOAA Physical Sciences Laboratory}},
  title = {CPC Global Temperature and Precipitation datasets},
  year = {2025},
  url = {https://downloads.psl.noaa.gov/Datasets/cpc_global_temp/Summary, https://downloads.psl.noaa.gov/Datasets/cpc_global_precip},
  note = {Accessed: 2025-05-15}
}

@data{j49q-ch56-25,
  doi = {10.21227/j49q-ch56},
  url = {https://dx.doi.org/10.21227/j49q-ch56},
  author = {Hengyuan He},
  publisher = {IEEE Dataport},
  title = {California Traffic Network Datasets: METR-LA, PEMS-BAY, PEMS04 and PEMS08 for Traffic Speed and Flow Analysis},
  year = {2025}
}

@inproceedings{koh2017influence,
  title={Understanding Black-box Predictions via Influence Functions},
  author={Koh, Pang Wei and Liang, Percy},
  booktitle={ICML},
  year={2017}
}

@article{aboitiz1992fiber,
  title={Fiber Composition of the Human Corpus Callosum},
  author={Aboitiz, Francisco and Scheibel, Arnold B. and Fisher, Robin S. and Zaidel, Eran},
  journal={Brain Research},
  volume={598},
  number={1--2},
  pages={143--153},
  year={1992}
}

@inproceedings{hamilton2017inductive,
  title     = {Inductive Representation Learning on Large Graphs},
  author    = {Hamilton, William L. and Ying, Rex and Leskovec, Jure},
  booktitle = {Advances in Neural Information Processing Systems (NeurIPS)},
  volume    = {30},
  year      = {2017}
}

@inproceedings{velickovic2018graph,
  title     = {Graph Attention Networks},
  author    = {Veli{\v{c}}kovi{\'c}, Petar and Cucurull, Guillem and Casanova, Arantxa and Romero, Adriana and Li{\`o}, Pietro and Bengio, Yoshua},
  booktitle = {International Conference on Learning Representations (ICLR)},
  year      = {2018}
}

@inproceedings{brody2022attentive,
  title     = {How Attentive are Graph Attention Networks?},
  author    = {Brody, Shaked and Alon, Uri and Yahav, Eran},
  booktitle = {International Conference on Learning Representations (ICLR)},
  year      = {2022}
}

\end{document}